# A Utility Maximization Model of Pedestrian and Driver Interactions

Yi-Shin Lin, Aravinda Ramakrishnan Srinivasan, Matteo Leonetti, Jac Billington, & Gustav Markkula

*Abstract*—Many models account for the traffic flow of road users but few take the details of local interactions into consideration and how they could deteriorate into safety-critical situations. Building on the concept of sensorimotor control, we develop a modeling framework applying the principles of utility maximization, motor primitives, and intermittent action decisions to account for the details of interactive behaviors among road users. The framework connects these principles to the decision theory and is applied to determine whether such an approach can reproduce the following phenomena: When two pedestrians travel on crossing paths, (a) their interaction is sensitive to initial asymmetries, and (b) based on which, they rapidly resolve collision conflict by adapting their behaviors. When a pedestrian crosses the road while facing an approaching car, (c) either road user yields to the other to resolve their conflict, akin to the pedestrian interaction, and (d) the outcome reveals a specific situational kinematics, associated with the nature of vehicle acceleration. We show that these phenomena emerge naturally from our modeling framework when the model can evolve its parameters as a consequence of the situations. We believe that the modeling framework and phenomenon-centered analysis offer promising tools to understand road user interactions. We conclude with a discussion on how the model can be instrumental in studying the safety-critical situations when including other variables in road-user interactions.

*Index Terms*—Adaptive control, agent-based modeling, behavior model, decision theory, intelligent vehicles, multi-agent system, utility theory.

## I. Introduction

Human locomotion, alone or in relation with others, is a long-standing topic of empirical and modeling research, not least in the context of road traffic [1]. Recently, the push towards increasingly fully automated vehicles has highlighted further the importance in this line of research. Crucially, one of the remaining challenges hindering automated vehicles from operating in urban environments is the lack of a sufficiently detailed understanding of how humans interact with others in such conditions, ideally operationalized in quantitative models [2], [3]. These models are needed both as components in online algorithms, to make real-time predictions about surrounding road user trajectories [4], [5] and as virtual agents in offline simulations, for safety testing of the algorithms driving automated vehicles before deployment [6], [7]. Automated vehicles, like human drivers, must trade between being sufficiently cautious to keep the risk of crashes minimal with being assertive enough to make meaningful progress in traffic. This balance is achieved in human drivers by being sensitive to subtle cues in the behavior of others, enabling rapid resolution of conflicts when needed [8], [9]. The balancing task is almost always successful in human drivers, but occasionally they run into safety-critical situations. To safely deploy automated vehicles in urban environment, they must resolve the safety-critical situations equivalence to or better than human drivers. Thus, the models of human behavior will be instrumental in understanding both the subtleties of local conflict resolutions between road users and the situations of unresolved conflict, resulting accidents.

To attain this objective, we propose a model that can account for local interaction by applying three basic principles - maximizing one's utility, actions as simple motor primitives, and triggering an action based on accumulated sensory evidence. Several previous works applied the utility maximization idea in accounting for road user interactions [10]–[13]. Here, we connect this principle to the sensorimotor control framework and evidence accumulation to make discrete action decisions. The utility maximization (UM) in the model, is situation dependent. Hence, when a particular situation arises, such as a collision conflict between two vehicle agents, the model adapts and applies the UM principle to account for the agents' interactive behaviors. The approach of applying basic principles mitigates the problem of growing model complexities by proposing the conceptual shift from merely quantifying every perceptual information to associating it with one's psychological utility. In this paper, we start from examining three situational variables, a) agents' preference in consuming energy, b) their desire to reach their destination and c) their emotional stress towards a potential collision. We detail these variables in Part II.

Most existing models of road user behavior have yet to meet the two above-mentioned criteria. Many models describe the interactions of a multitude of road users; thus allowing for predictions of traffic flow at the infrastructure level [14], [15],

Manuscript received xxxx xx, 2021; This work was supported by EPSRC under Grant number, EP/S005056/1. (*Corresponding author*: Gustav Markkula and Yi-Shin Lin.)

Yi-Shin Lin is with the Institute for Transport Studies, University of Leeds, LS2 9JT UK (email: Y.S.Lin@leeds.ac.uk).
Aravinda R. Srinivasan is with the Institute for Transport Studies, University of Leeds, LS2 9JT UK (email: A.R.Srinivasan@leeds.ac.uk).
Jac Billington is with the School of Psychology, University of Leeds, Leeds, LS2 9JT UK (e-mail: J.Billington@leeds.ac.uk).
Matteo Leonetti is with the School of Computing, King's College London, London, WC2R 2LS UK (e-mail: matteo.leonetti@kcl.ac.uk).
Gustav Markkula is with the Institute for Transport Studies, University of Leeds, LS2 9JT UK (e-mail: G.Markkula@leeds.ac.uk).

and also permitting simulations of crashes caused by human error [16], [17], but they seldom emphasize local interactions which are keys in critical situations. Further, models which do examine local interactions among road users, mostly assume that agents move to keep in lane, maneuver around obstacles and so on [18]–[20] and that surrounding agents are static obstacles, assuming no interactive conflicts and thereby no safety-critical situations may arise. The few models which have focused on the details of routine local-interactions between two or more road users, were designed for specific scenarios such as highway merging [21], pedestrian road-crossing [22], or shared-space locomotion [23]. Therefore, it is unclear how one can extend these models to other scenarios or safety-critical interactions. Machine-learning models could theoretically apply to a wide ranges of scenarios and have been tried on more complex interactions [24], [25]. Their very strength, however, depends considerably on the increasing amount of data and the extrapolation of these models to scenarios that are rare in naturalistic datasets, such as safety-critical situations, may not entirely reliable. Thus, the data-driven models are not optimal for determining and more importantly, understanding how humans resolve interactive subtleties [26].

Here, we leverage a cognitive theory for modeling sensorimotor control [27], which describes movement as consisting of discrete action primitives, triggered after one gathers sufficient sensory information favoring a chosen action. This framework has been utilized in modeling driver behavior, investigating both near-optimal, closed-loop movement control (as is typical of routine driving) and sub-optimal open-loop maneuvers triggered after long, uncontrolled delays (as is typical of near-crash driving). This framework has been shown capable of accounting for driver behavior across a range of situations including both routine and near-crash driving [27]–[31].

Recently, we and others have begun expanding this modeling approach to interactive traffic situations. Boda and colleagues [32] showed that the framework could be extended to account for how drivers control longitudinal maneuvers when cyclists crossed the drivers' path. Others researchers had shown the onset latency of the road-crossing decisions in drivers and pedestrians could be modeled by the principle of gathering sensory cues [33]–[35]. However, a more generic framework, valid across different road user types and scenarios, is still lacking. This paper presents a first step in this direction, extending the framework to model generic road-user agents and applying the framework to interactive, time-evolving scenarios.

One modeling decision that we make is to formulate the framework, using the UM principle. Our previous modeling work had instead emphasized using the association between perceptual quantities (e.g., optical invariants [19]) and motor primitives for deciding on actions (i.e., action decisions), but perceptual quantities tend to be highly situation-specific, and we hope applying the basic principle of utility maximization may generalize the model formulation.

When aiming to account for complex behaviors, the models themselves necessarily become complex, with a rapidly branching tree of modeling decisions, and therefore a primary concern becomes: Which model assumptions are needed to capture what behavioral phenomena? Instead of starting with investigating individual road-user behaviors, we take advantage of the computation power of modern computers by surveying many candidate models and examining them against a selection of typical road-user behaviors reported in literature. This method is to study the parameter space and to find the candidate models belonging to certain parameter regions can manifest the typical road-user behaviors qualitatively. This approach resemble the *parameter space partitioning* used by cognitive modelers [36]–[38], but has seen limited use in the road user modeling domain.

Using this approach, we show that our framework can naturally account for several empirical phenomena that have been previously reported in interactions between pedestrians and between pedestrians and drivers, and in each case, we also provide a more detailed demonstration of how the model reproduces the road user behaviors in question, by comparing to empirical data from the literature.

As a first step in this endeavor, we start with simple routine interactions, where the interacting agents are on straight crossing paths. In this paper, we first report deterministic model simulations, show how the deterministic simulations can further incorporate stochastic component to show human-like behavioral patterns, and then study the influences of time gaps on traffic decision-making.

Section II introduces the modeling framework, two target interactive scenarios and how we quantify behavioral data to test the model in these scenarios. Section III presents the results of the model analyses, and Sections IV and V discuss the results, focusing on how the principle of utility maximization and its derived model can be extended to study other kinematic variables, such as the influence of perceptual assessment of the vehicle acceleration on decision-making, how one may incorporate the associated kinematic variables to study road user interactions and how this may play a critical role when the situations deteriorate. We conclude that the three basic principles enable the framework to be readily extendable and to account for further variables in the road user interactions both in normal and safety-critical situations.

## II. METHODOLOGY

### A. Modeling Framework

In this section, we describe the details of how the UM model can account for road user interactions, how it can apply to two interactive scenarios between pedestrians and drivers, and the core tenets of the framework: motor primitives, the utility function, and the stochastic action decisions.

*1) Motor Primitives*

The sensorimotor control framework [27] assumes agents move by initiating an intermittent, discrete, and ballistic motor primitive [39]. The motor primitive acts as a limited-duration motoric adjustment to an ongoing motion. The motoric adjustment cannot be changed once initiated but can act atop any ongoing adjustments, for example allowing midway changes of action before another primitive completes.

We assume the agent possesses a limited discrete set of motor primitives, with vehicle agents adjusting acceleration and yaw rate, and pedestrian agents speed and yaw angle. In the simulations below, we consider only longitudinal movement; therefore, any changes in yaw rates or angles become inconsequential. This simplifying assumption, as shown in the Result section, is applicable to a suite of data from both laboratory and field studies. Furthermore, we assume vehicle agents possessing the repertoire of motor primitives, -1, -0.5, 0, 0.5, and 1 m/s$^2$ and pedestrian agents, -1, -0.5, 0, 0.5, and 1 m/s. These primitives adjust the ongoing action. All primitives are implemented as a constant pulse of signal of a duration $\Delta T$, here set to 0.3 s. For example, if pedestrian agents at standstill decide to apply a motor primitive of 1 m/s, their speed will increase linearly from zero to 1 m/s over $\Delta T$ seconds, and then remain at 1 m/s. While that adjustment is taking place the agent could decide to apply another motor primitive that acts on top of the ongoing action.

*2) Agent's Utility Function*

At each step, the agent foresees $T_p$ seconds to assess what states each motor primitive will lead to, assuming other agents remain in their current kinematic states (see the Discussion about this assumption). The agent then calculates the utility corresponding to each of the primitives and favors the one resulting in the maximum utility at that step. Here, we assume the action duration is equivalent to that of the prediction ($T_p = \Delta T$). The agent associates a future state with a utility prediction, assuming the following utility function:

$$U = -k_g \dot{d}_g - k_{dv} v^2 - k_{da} a^2 - \sum_{i=1}^{n} C_i \quad (1)$$

The velocity and acceleration are denoted by, $v$ and $a$; $\dot{d}_g$ denotes the rate of change of the distance $d_g$ to a goal. The agent, in this paper, engages only in longitudinal control, so $\dot{d}_g$ is further simplified as $-v$, reflecting an agent's desire to reach its destination. The agent assumes the velocity and the acceleration components as energy consumption in the vehicle agent and as discomfort in experiencing high speeds and rapid accelerations in the pedestrian agent. Such assumption is common in the models of minimizing cost in sensorimotor control [40], [41][1]. The agent assumes a collision discomfort term, $C_i$, when it is on a collision course with other agents or obstacles $i$. In this paper we only consider a single other agent, therefore, $n = 1$. If a motor primitive does not predict a collision course with others, $C_i$ reduces to zero, whereas if it does predict such a possibility, identified as the two agents or an agent with an obstacle coming within a tolerance distance $d_c$ [2], then for the constant-speed, speed-control pedestrian agent, $C_i$ becomes $k_c \frac{1}{\tau_i}$, where $\tau_i$ is the time left to collision from a predicted state. This is a common assumption in the literature examining the role of the inverse time to collision in locomotion [42], [43]. The acceleration-control vehicle agent assumes this term as $C_i = k_{sc} a_{sc,i}^2$, where $a_{sc,i} = v^2/2d_i = v/2\tau_i$, the acceleration needed to stop before reaching the collision point a distance $d_i$ away [44]. Theoretically, by partial differentiating (1) regarding to $v$ and assuming $C_i = 0$, the maximum utility is achieved when agents are traveling at their theoretical free speed $v_{free} = k_g/2k_{dv}$.

*3) Action Decisions*

An action decision is only initiated when its utility gains a positive margin over the others. In every time step, a favored action is derived by sweeping through simultaneously all motor primitives, $m = 0, 1, \cdots, n$, each corresponding to a situation utility, $U_m$, determined both by the agent characteristics, operationalized as the model parameters and by the situational variables. This sweep includes $U_0$, the utility of not applying any motor primitive. If one or more motor primitives $m \neq 0$ is identified with $\Delta U_m \equiv U_m - U_0 > 0$, the motor primitive with the highest $\Delta U_m$ is selected, and the corresponding action is then initiated.

To extend the framework to probabilistic behavior, we include noisy evidence accumulation using a standard formulation [27] as follows:

$$\widehat{\Delta U}_m(k) = \widehat{\Delta U}_m(k-1) + \frac{d\widehat{\Delta U}_m(k)}{dt}\Delta t + \sigma \epsilon(k)\sqrt{\Delta t} \quad (2)$$

where $\widehat{\Delta U}_m(k)$ is the accumulated evidence in favor of motor primitive $m$ at time step $k$ of a simulation with time step length $\Delta t$, and where $\sigma \epsilon(k)$ is Gaussian noise with standard deviation $\sigma$. The agent gathers evidence up to a threshold, $\widehat{\Delta U}_m > E_a$, at which point the motor primitive $m$ is triggered, and all $\widehat{\Delta U}_m$ are reset to zero. The rate of change of the evidence could be implemented in a number of different ways; here we have opted for a formulation which makes $\widehat{\Delta U}_m$ interpretable as a low-pass filtered version of a noisy $\Delta U_m$:

$$\frac{d\widehat{\Delta U}_m(k)}{dt} = \frac{1}{T}\left(\Delta U_m(k) - \widehat{\Delta U}_m(k-1)\right) \quad (3)$$

where $T$ is a scaling parameter on the rate of evidence growth. Thus, we transform the idea of evidence accumulation, proposed for instance in [30], to not only enable the agent to travel but also motivate it to maximize the utility. A specific treatment, "squashing" [45] is applied to $\tau_i$-associated utility, constraining it from growing to negative infinity.

*B. Interactive conflicts and manifested behaviors*

In this section, we first divide the interactions into two categories, pedestrian-pedestrian and pedestrian-vehicle interactions. In each category, we describe the assumptions that facilitate the investigation of each interactive conflicts, and the modeling set-ups that enables the examination of the parameter regions generating the natural behaviors. Second, we operationalize the natural behaviors generated by the model simulations, based on the literature reporting these behaviors. Third, we describe the experiment designs, assumptions and metrics used in the previous studies [33], [36]–[38], [42], [46], [47] and the model set-ups used to reproduce the qualitative patterns observed in these studies.

*1) Pedestrian-pedestrian conflict*

When two pedestrians are crossing each other's paths, they naturally adapt to avoid collision. This was studied in a pair of publications by Olivier et al., where each of the two human participants walked along crossing diagonal lines in a square-shaped room where a curtain initially occluded them from seeing each other, as illustrated in the upper panel in fig. 1 [46], [48]. A main finding was that the pedestrian who was closer in time to the prospective collision zone at the time $t_{see}$ tended to speed up and pass the midpoint first, and the other participant contributed to the collision resolution accordingly by decelerating to give way. $t_{see}$ refers to the time when two pedestrians were first able to see each other. These behaviors imply that the participants negotiated the priority without verbal communication but by observing the other's kinematics state.

---

[1] An automated vehicle likely manifests very different tolerance in accelerating than a human driver.

[2] Assuming the agents maintains their state.

We assumed two homogeneous pedestrians by assigning them identical model parameters and symmetrical initial positions to simulate the pedestrian-conflict paradigm. To create asymmetric kinematics at $t_{see}$, we assigned the agents with different initial speeds. In this first scenario, we investigated the parameter regions of the discomfort of speeding up, $k_{dv}$ and of colliding, $k_c$, with an aim to reproduce the empirical observations [46], [48]. We fixed the parameter $k_g$ reflecting the desire of reaching the destination at 1, without loss of generality since the utility function is in arbitrary units (all parameters can be scaled by an arbitrary constant without changing the relative utilities of different actions), and for simplicity, when not noted, we assumed agents are free of acceleration discomfort, $k_{da} = 0$. We then tested different combinations of $k_{dv}$, and $k_c$, with the former restricted to the range of 0.28 to 0.71 and the latter to the range of 0 to 10. The region was selected because they produce pedestrian free walking speed, $v_{free}$ 0.7 and 2 m/s, which was often observed in urban and suburban environment [49]. We varied the agents' initial speeds from 0 to 0.9 m/s with a step size of 0.1 m/s. In total, for each model parameterization ($k_{dv}$, $k_c$) we tested 100 different pairs of initial speeds.

We summarize the key observations in [46], [48] and define them as the following Boolean metrics to investigate the model simulations: (1) *Lead agent passed first (LAPF)* is true, if the lead agent passes the crossing point earlier than the other agent. The lead agent is the one with the shortest time to the crossing point at $t_{see}$, assuming constant speeds; (2) *First-passer accelerated (FPA)* is true, if the lead agent increases its speed just after $t_{see}$, operationalized as a speed $v(t_{see} + T_a) > v(t_{see})$, with $T_a = 0.5$ second; (3) *Second-passer decelerated (SPD)* is true, if the lag agent decreases its speed just after $t_{see}$.

We consider a model parameterization effectively reproducing the natural behaviors if the LAPF is true in more than 80 %[3] among the 100 simulations of the different initial speeds and if the FPA and SPD are both true in more than 20 %[4]. We set these somehow arbitrary criteria on the basis of inferring from the previous data [46], which reported the walkers in lead position almost always passed first or accelerated to pass first when at a slower speed at the $t_{see}$. A slight change in the percentages, for instance changing LAPF to 75 % or to 85 % or changing FPA to 15 %, does not alter the qualitative pattern of the result. The main purpose of using these criteria is to investigate regions in the parameter space that produce the qualitative patterns in the natural data [36]–[38]. The result in fig 4 supports the usefulness of these criteria.

Moreover, we define three excluding metrics, to discard model parameterizations yielding behaviors that were in direct conflict with the natural behaviors: (1) *First-passer decelerated* (FPD) (2) *Second-passer accelerated* (SPA), and (3) *Collision occurred* (CO). The first two are defined analogously to FPA and SPD above and the latter one is true if the agents collided (distance below $d_c = 1$ m). We consider the model parameterizations reproducing unnatural behaviors when any of the three metrics exceeds 5%.

In addition to the excluding metrics, we compare the resulting trajectories from our simulations quantitatively with the observations in [46], by calculating the *minimal predicted distance* (MPD) metric, $d_{min}$. The metric was defined in [46] as the smallest future distance at any point in time $t$ between the agents, assuming that they will travel in constant speeds from $t$ onward.

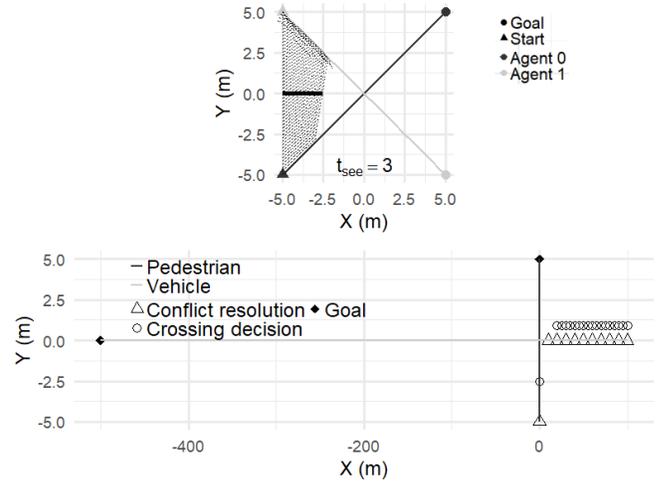

Fig. 1. Two interactive scenarios, depicted from a top-view perspective. In the upper panel, the dotted lines show from the line-of-sight of Agent 0 to Agent 1. $t_{see}$ represents the time when the agents first see each other behind the dividing curtain (the horizontal thick black line). The lower panel shows the paths of a pedestrian and a vehicle. The triangle and circle represent different starting positions in the two scenarios, *conflict resolution* and *crossing decision*. The open symbols show different starting positions. The solid diamond show the agent's goal destinations. We avoid clutter by drawing the starting positions in *Crossing Decision* scenario off the axes.

*2) Pedestrian-vehicle conflict*

Typical observed time gaps in which pedestrians make attempts to cross roads when a vehicle drives toward the crossing are between 2 to 6 seconds [42]. Specifically, the time gap refers to the moment-to-moment difference in the time to the crossing point between the two agents. Comparing to the interaction in pedestrians, the interaction between pedestrians and vehicles often manifests a different set of behaviors, for instance, the lead agent might not pass first [47]. Importantly, the time gap influences these behaviors. The accelerating agents, like vehicles, often asserted the lead role to pass earlier even when they were not in a lead position [47], [50].

In the second part of the investigation in the road-user interactive conflicts, we again harness the modeling framework to examine two questions: (1) how the time gap affects *crossing decisions* (CD) and (2) how pedestrians and drivers reach *conflict resolution* (CR). We define the natural behavior of time-gap dependent CD as: When the time gaps are small, the pedestrian yields to the vehicle, whereas, when the time gaps are large, the pedestrian may cross in front of the vehicle. In between, the crossing decisions depend on pedestrians' evaluation of the vehicle kinematics. The parameter, $k_c$ reflects such evaluations. CR, on the other hand, was observed as the road user noticeably alter their behavior either to slow down or to speed up [8], [22]. These CR actions are often read as gauging the uncertainty in decelerations, or as asserting priority in accelerations. Therefore, we define the natural behavior of CR in the pedestrian as yielding in the "encounter" (see the following for its definition) time gaps.

---

[3] 100 % is not to be expected, because the times to crossing at $t_{see}$ will sometimes be nearly or completely identical.

[4] The FPA and SPD are expected to be considerably lower than 100 %, because for many combinations of initial speed, most cases the lead agent was in a higher speed than that of the other agent.

To examine these two sets of natural behaviors, we set up two scenarios to investigate how the time gaps associate with the model parameters and how they associate with the CD and CR behaviors. The first scenario, denoted CD, places a vehicle at 17 different locations on the x axis, starting from 20 m to 100 m by a 5-m step. These starting positions result in 17 time gaps when the vehicle starts at its free-flow speed at 13.9 m/s (~ 50 km/h). Meanwhile, a pedestrian stands still on the negative side of the y axis at 2.5 m. The second scenario, denoted CR, places a free-speed vehicle at ten different starting positions, from 10 m to 100 m by a 10-m step on the x axis. The pedestrian in this scenario walks at a speed of 1.1 m/s, starting from the negative side of the y axis at 5 m away from the crossing point. One major difference between the two scenarios is the pedestrian starts at zero speed in the former and, in the latter starts at a walking speed.

These two scenarios focus on the behavior of collision aversion and thus examine how the model parameters, $k_c$ in the pedestrian and $k_{sc}$ in the vehicle jointly affect a pedestrian to make crossing decisions and to resolve collision conflicts. The two parameters were selected because how different road users – pedestrians, human drivers, and automated drivers – perceive and weight the risk of collision has significantly societal impact both on current and near-future traffic.

Parallel to the approach in the pedestrian interaction, we systematically examine 100 evenly spaced $k_c$ and $k_{sc}$ values from 0 to 5, across the time gaps in CD scenario. We reduce the upper limit to 2.5 in the CR scenario and keep surveying 100 evenly spaced parameters, because in the deterministic case, the values above 2.5 did not change the result substantially.

After showing the CD result in the first scenario reproduces the time-gap data [42], we carried on investigating whether the model can reproduce the CD data [33], showing an association between the response time distribution and the time gaps. We reuse data from the study [33], which invited 20 human participants to perform a road-crossing task in a virtual reality environment where a constant-speed vehicle drove towards the crossing zone. The study divided the data set into three groups based on the time gaps, 2.29, 4.58 and 6.87 s [33].

Similarly, the parameter survey in the second scenario was followed by an investigation to see whether the utility-maximization model is capable of reproduce the CR data in a field study [47]. The drivers in this study accelerated to assert their priority, even when they were in a lag position. Following [47], we define an "encounter" as a situation, which a pedestrian and a vehicle would reach the crossing point within 1 s of each other's prediction time, assuming they maintain their speeds. The encounter time then includes both the cases when either the vehicle or the pedestrian is closer to the crossing point (i.e., lead position). We create 859 "encounter" time gaps in this scenario by placing a pedestrian agent on 60 evenly spaced positions between -10 m to -5 m on the y axis and a vehicle agent also on 60 evenly spaced positions between 10 m to 79 m. The two agents assumed their speeds at 1.4 m/s and 13.9 m/s similar to those reported in [47] and reacted to each other immediately after the simulation started.

It was less likely that the assertive behavior in the drivers was due to most of them being prone to take risks. Instead, we assume their assertive behavior is due to the accelerative nature of the vehicle and thus we rendered the vehicle agent in this simulation less sensitive to speed discomfort by decreasing $k_{dv}$. As a result, the vehicle agent elevates its free speed and becomes more liberal to accelerate in response to other agents.

III. RESULTS

A. *Pedestrian-pedestrian conflict*

Fig. 2 summarizes the parameters producing the natural behavior of pedestrian conflict resolution in the "accepted" region, which roughly enveloped two strips with some exceptions. The summary suggests the natural behaviors in pedestrian manifest a tendency to avert collision. That is, the parameters with near-zero $k_c$, near-zero $k_{dv}$ and large $k_{dv}$ have not produced natural behaviors. Fig. 3 shows the results for each criterion. Almost all surveyed parameters could reproduce LAPF irrespective of the initial speeds. The two associated behaviors, the FPA, and the SPD, filtered out parameters mostly, when $k_c$ values were near 0, reflecting that the pedestrian rarely ignores collisions completely, and when $k_{dv}$ values were very large, reflecting that the pedestrian does not increase speed drastically. The contrasting behaviors, FPD, and SPA, excluded a good number of parameters, as these two criteria describe atypical pedestrian behaviors. In rare occasion when $k_{dv}$ values were small, the parameters resulted in collisions.

Fig. 4 is the result of the model simulations compared to the empirical data. We generated the trajectories using the 3072 "accepted" parameters in fig 2 and then calculated their MPD between the agents at every time point, before they reached the intersection. These trajectories were then divided into ten groups, according to an ascending order of MPD at the time of $t_{see}$, as in Olivier and colleagues' fig. 4 [46]. The model simulations show when the MPDs were ordered by the $t_{see}$, the predicted MPDs at the $t_{cross}$ are in line with the empirical data.

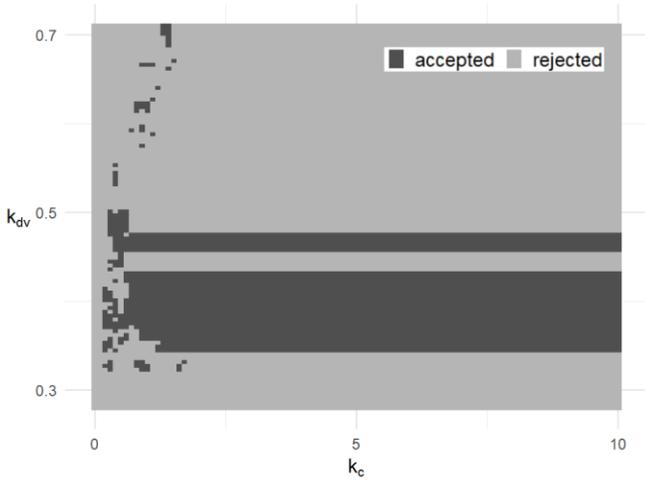

Fig. 2. The parameters reproducing the behavior of leading agent passing first.

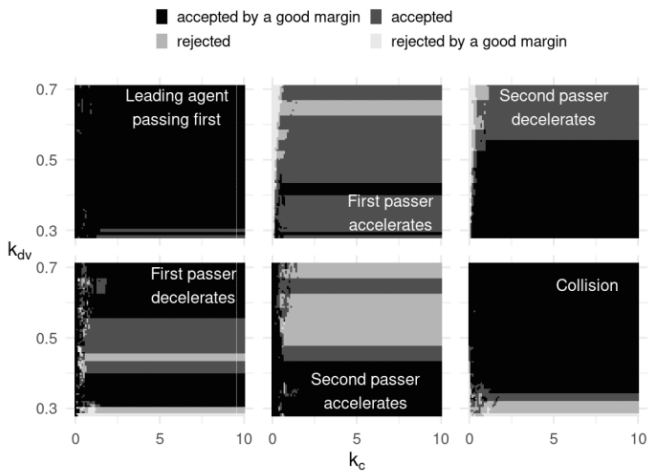

Fig. 3. Individual criterion on the parameters reproducing the typical behavior of pedestrian interaction.

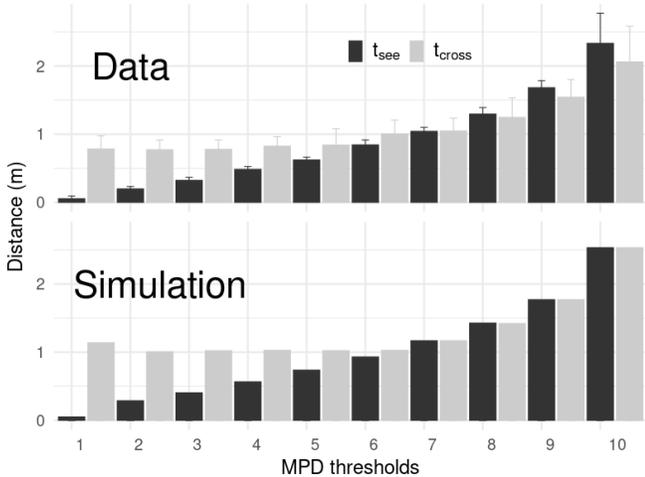

Fig. 4. Comparison of minimal predicted distances between the empirical data in [46] and the simulations, at both $t_{see}$ and $t_{cross}$.

### B. Pedestrian-vehicle conflict
#### 1) Crossing decision

Fig. 5 shows the evolution of speeds, accelerations, and the distance to the crossing in three simulation examples, where a pedestrian reacted to a constant-speed vehicle. When the time gap was 2.16 s, the pedestrian decelerated to yield, when the time gap was 4.32 s, the pedestrian passed earlier than the vehicle by stepping up the speed to 1.5 m/s, and when the time gap was 6.47 s, the pedestrian passed without accelerating drastically. Fig. 6 summarizes the results of the first scenario, showing the minimal time gap for three selected $k_{sc}$ values. When a pedestrian was cautious ($k_c = 0.051$) and the vehicle paid little attention to others ($k_{sc} = 0$), the shortest time gap a pedestrian could cross safely was around 3.96 s. The time gap decreased to 2.88 s when the vehicle was moderately cautious ($k_{sc} = 0.51$), yielding to even pedestrian paying little attention to avoid collision (i.e., $k_c = 0$). The time gap decreased further to 2.16 s when the vehicle was very cautious ($k_{sc} = 1.01$). The result in $k_{sc} = 0$ reproduces the time gaps reported in Experiment 2 in [42], where the participants in the age groups of 20-30 and 60-70 made crossing decision between 3 to 4 s.

In addition, the utility-maximization model can be rendered to reproduce the response times in pedestrians when they reacted to different time gaps. We choose the parameters, $k_g = 1$, $k_c = 1$, $k_{dv} = 0.38$ based on the parameter survey in deterministic simulations and apply (2) and (3) to generate noisy behaviors of pedestrian-vehicle interaction, shown in fig. Fig. 7. The model is capable of catching the critical information even in a small data set giving only coarse response distributions [33]. Firstly, when the time gap was 2.29 s, the data showed the participants entered most of their crossing responses after the vehicle had passed, so does the model. Secondly, when the time gap was long, namely 6.87 s, the data showed substantial crossing responses were entered before the vehicle passed. So does the model. Third, when the time gap was in between, the data showed a balance of pre- and post-vehicle pass responses. The model in this case predicted a small proportion of post-vehicle responses. In summary, the model has shown a good capacity to account for both the field- and laboratory-observed data in crossing decisions.

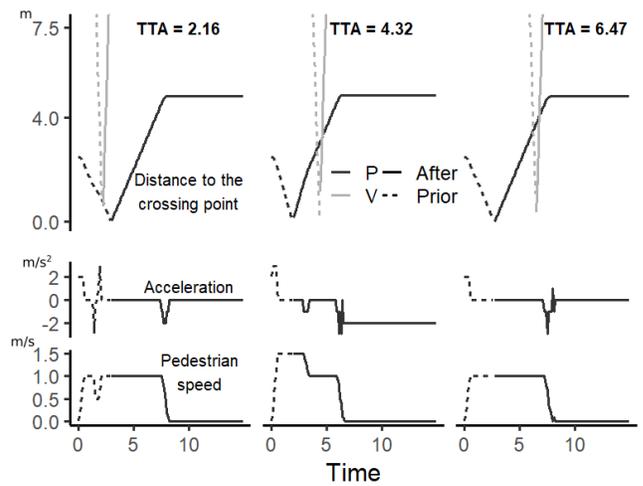

Fig. 5. The behavior of crossing decision in three time gaps. The parameters were extracted from the top panel in fig 6 where no collision occurred.

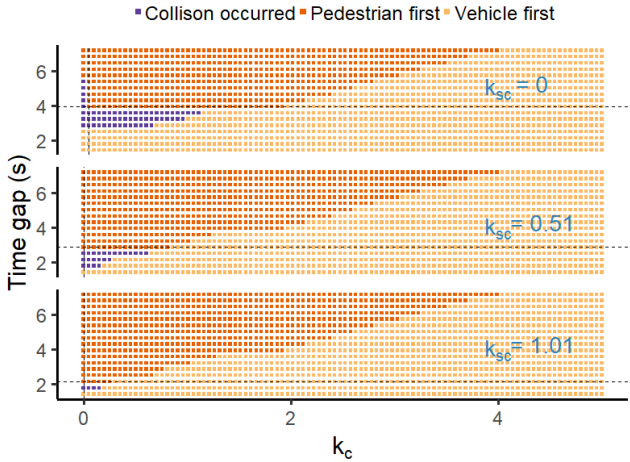

Fig. 6. The association between the time gaps and the collision aversion in the pedestrian and the vehicle agent. The crosshairs highlight the time gaps described in text.

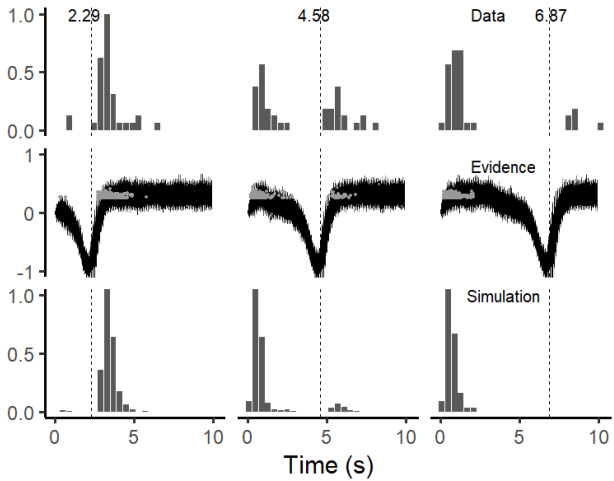

Fig. 7. Model simulations of stochastic road-crossing decisions. The time gaps are highlighted by the dashed lines. The black traces in the middle row are the evolution of evidence values, and the grey dots are the resultant responses. The upper and lower rows show the density histogram of the response times.

*2) Conflict resolution*

Fig. 8 summarizes the result of the second scenario, surveying the region of the two collision-discomfort parameters, $k_{sc}$ and $k_c$. The low-left corners in the fourth panel showed neither agent yielded. These are the parameters enabling the agents to be more willing to take risks (i.e., $k_c < 0.28$ & $k_{sc} < 0.13$). Of course, a risk insensitive road user does not always run into the other, cautious, road user. For example, in the first panel, when the pedestrian was prone to risk-taking ($k_c < 0.28$), but the vehicle was rendered to be risk-averse, the vehicle yielded. Overall, the pedestrian and vehicle agents yielded when their collision-discomfort parameters, $k_c$ and $k_{sc}$, are greater than 0.28 and 0.38, although this depends on the time gaps.

The CR simulation for reproducing drivers' assertive behavior results in 72% of the vehicle agents accelerated in the "encounter" time gap and 86% of them were observed when the vehicle was in a lag position. This result in line with what the field study observed, where 73% of the driver maintained the same speed or accelerated [47].

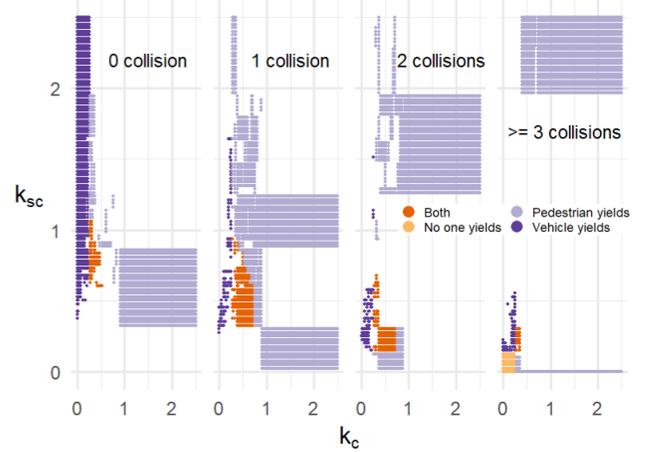

Fig. 8. The parameter survey of conflict resolution. The number of collisions was tallied in the ten tested time gaps.

## IV. DISCUSSION

We developed a modeling framework by applying the basic principle of maximizing one's psychological utility in order to understand interactions among road users in traffic. By applying the basic principles - motor primitives, utility maximization, and intermittent action decisions - our model enables a principle-based framework to successfully reproduce different scenarios of traffic interactions, where road users resolve interactive conflicts and make critical decisions. The model, in its current form, has been instrumental to discern the time-gap influences on pedestrians' road-crossing decisions.

Many previous works focused on accounting for the perceptual and attentional aspects in processing of sensory information with the traffic tasks in mind [51], [52]. These models resulted in fruitful quantitative descriptions in the traffic task performance but depend on perceptual quantities that must be painstakingly identified for each new scenario. Instead of continuing to focus on accounting for many perceptual differences in traffic tasks, in this paper, we applied the basic principle of utility maximization to account for deterministic interactions and then showed how the deterministic interactions can be extended to stochastic road-crossing decisions by associating the utility values with the rates of gathering sensory evidence for action decisions. The approach of applying basic principles mitigates the problem of growing model complexity by proposing the conceptual shift from merely quantifying every perceptual information to associating it with one's psychological utility.

Specifically, we started from the existing framework for modeling sustained sensorimotor control [27], previously mainly applied to modeling of non-interactive control behavior of single drivers, and generalized it to model also road user interactions, including both driver and pedestrian agents. The rationale for building on this specific framework is that it has been shown capable of accounting for not only routine driving control, but also near-crash behavior, including probability distributions of open-loop defensive reactions, which are enabled by the evidence accumulation and motor primitive assumptions of the framework [27]–[31] – notably the same assumptions which here allowed us to model entire probability distributions of pedestrian crossing onset (fig. Fig. 7). An important avenue for future work will be to leverage these capabilities of the framework to study and model near-crash behavior also in multi-agent interactions. One requirement for doing so is likely to be a further extension of the framework to model agent beliefs about the intentions of other agents, since this is an important cause of safety-critical interaction failures [53], [54]. Such an extension would also be beneficial to further improve the model's account of also non-critical interactions, since both empirical observation [3], [9], [35] and everyday experience clearly suggests that belief about others' intentions play an important role in many of the subtler details of local road user interactions. As mentioned in the introduction, computational models that capture these subtleties are needed to help enable safe and acceptable vehicle automation [2], [51]. For first indications of how intention beliefs can be incorporated in the present framework, see [31], [35].

Here, we started by testing model candidates that generate natural behaviors and studied on parameter space to understand the ability of the model for accommodating the behavior in question between pedestrians and between a pedestrian and a vehicle. An immediate next step is to study the concept of utility maximization in the interactions between drivers, where we found, using the neural network of convolutional social pooling, the drivers in the lead position stepped up speed to assert their way when driving on a merging lane [26]. Although it remains an open question as to whether the UM principle can be applied in the field data extracted from uncontrolled, video-recorded data, we believe the detailed behavior of driver interactions could also be accounted for by applying the principle.

Although, in this paper, we focused only on the influence of time gap on crossing decisions, there are four other variables that might play critical roles in road users' crossing decisions. These are the distance gap, the perception of the yielding deceleration, the size of the approaching agent and explicit signals of communication. The capacity of the model can be further extended by adding these situation variables and assuming their corresponding model parameters. Next, we sketch the plan to investigate these variables and how the UM model may help to understand their roles in traffic interactions.

Although the time gap, distance, and the speeds are bounded by classical mechanics, they appear not always to be processed in strict physics principle by human road users. For example, to make a left turn maneuver, a driver seemed to rely more on the perceived distance to an opposing vehicle than on the perceived speed and the time gap [55]. In a similar example, drivers of over 56 years old accepted the same distance gap, irrespective of whether an approaching vehicle was at different speeds [56]. These findings imply that the perceived distance and decisions thereof depend not only on the physical but also on the psychological factors. Second, whether an observation of deceleration is read as a yielding sign also affects the crossing decisions. A study, analyzing a series of traffic videos, showed the clear sign of a vehicle to decelerate associated with an increase in the probabilities of pedestrians' crossing decision [22]. Third, smaller vehicles, for instance cars versus motorcycles, are perceived as approaching later. This cognitive illusion, dubbed size-arrival effect, resulted in an increase in the risk of dangerous crossing decision in pedestrians when a mixture of difference vehicles are in traffic [57]. A similar observation was also reported, comparing trucks with motorcycles, even when both arrived at the same time [58]. These variables can be also tested in the UM model, which will help the understanding in how these variables affect the crossing decisions. A further step could be to examine how these many traffic-related variables contribute to human road-crossing decisions when, for instance, machine-driven vehicles are on the road. We expect these understandings will help to design a safer environment when human and autonomous vehicles coexist in future traffic.

One key issue is that the time resolution typically applied in in traffic simulation is vastly different from the time resolution often used in the orthogonal decision theory, such as the diffusion decision model [59]. In the former, to predict meaningful traffic flow, 0.1 s is the usual lower bound; however, in the latter, the decision process in humans is often examined in the split of 1 ms. This difference in time resolution makes bridging the traveling path of an agent to its decision point challenging. At this point, our model errs on the side of coarse time resolution to enable the agent to travel with natural progress as in real traffic. However, in some real critical incidents, a fatal decision happens in the range of tens to hundreds of milliseconds, as observed in many human decision-making tasks. We hope the next development of the UM model could enable such dynamic changes in time resolutions as in what might happen in real life traffic scenarios.

## V. Conclusion

To conclude, utility maximization is a promising conceptual principle, and the models derived from it provide a useful tool to describe traffic interactions in local details. We believe this concept can apply to examine more diverse traffic interactions, such as multiple vehicles agents resolving collision conflicts in highways. We hope the concepts, the modeling framework, and the software reported in this paper will inspire other researchers to take advantage of basic principle of maximizing utilities in studying traffic interactions.

## Appendix

We deposited the data and codes used in this paper in the OSF site, https://osf.io/av3cm/.